\ifdefined\pdfoutput\pdfoutput=1\fi

\documentclass[11pt]{article}

\usepackage{acl}

\usepackage{times}
\usepackage{latexsym}

\usepackage[T1]{fontenc}

\usepackage{microtype}

\usepackage[utf8]{inputenc}

\usepackage{graphicx}
\usepackage{booktabs} 
\usepackage{multirow} 
\usepackage{amsmath}
\usepackage{amssymb}
\usepackage{microtype}
\usepackage{graphicx}
\usepackage{subcaption}

\title{Enhancing Reasoning Abilities of Small LLMs with Cognitive Alignment}

\author{Wenrui Cai$^{1,2}$\thanks{\ \ The work was conducted during the internship at Alibaba Cloud Computing.}, Chengyu Wang$^2$\thanks{\ \ Corresponding author.}, Junbing Yan$^2$, Jun Huang$^2$, Xiangzhong Fang$^1$\\
  $^1$ Shanghai Jiao Tong University, Shanghai, China\\
  $^2$ Alibaba Cloud Computing, Hangzhou, China\\
  \texttt{\{cwrcwr,xzfang\}@sjtu.edu.cn}\\
  \texttt{\{chengyu.wcy,yanjunbing.yjb,huangjun.hj\}@alibaba-inc.com}}

\begin{document}
\maketitle
\begin{abstract}
The reasoning capabilities of large reasoning models (LRMs), such as OpenAI's o1 and DeepSeek-R1, have seen substantial advancements through deep thinking. However, these enhancements come with significant resource demands, underscoring the need for training effective small reasoning models. A critical challenge is that small models possess different reasoning capacities and cognitive trajectories compared with their larger counterparts. Hence, directly distilling chain-of-thought (CoT) rationales from large LRMs to smaller ones can sometimes be ineffective and often requires a substantial amount of annotated data. In this paper, we first introduce a novel Critique-Rethink-Verify (CRV) system, designed for training smaller yet powerful LRMs. Our CRV system consists of multiple LLM agents, each specializing in unique tasks: (i) critiquing the CoT rationales according to the cognitive capabilities of smaller models, (ii) rethinking and refining these CoTs based on the critiques, and (iii) verifying the correctness of the refined results. Building on the CRV system, we further propose the Cognitive Preference Optimization (CogPO) algorithm to continuously enhance the reasoning abilities of smaller models by aligning their reasoning processes with their cognitive capacities. Comprehensive evaluations on challenging reasoning benchmarks demonstrate the efficacy of our CRV+CogPO framework, which outperforms other methods by a large margin.\footnote{Source code is released in the EasyDistill toolkit~\cite{DBLP:journals/corr/abs-2505-20888}. URL: \url{https://github.com/modelscope/easydistill}}
\end{abstract}

\section{Introduction}

The remarkable progress in language reasoning models (LRMs) has revolutionized NLP~\cite{DBLP:journals/corr/abs-2303-18223}. Recently, leading models such as OpenAI's o1\footnote{\url{https://openai.com/o1/}} and DeepSeek-R1~\cite{deepseekr1} have leveraged slow, deliberative thinking to solve complex tasks. Despite their impressive capabilities, the scale of these models results in substantial computational demands. Consequently, there is a growing need to train reasoning models with fewer parameters.

\begin{figure}
\centering
\includegraphics[width=.95\columnwidth]{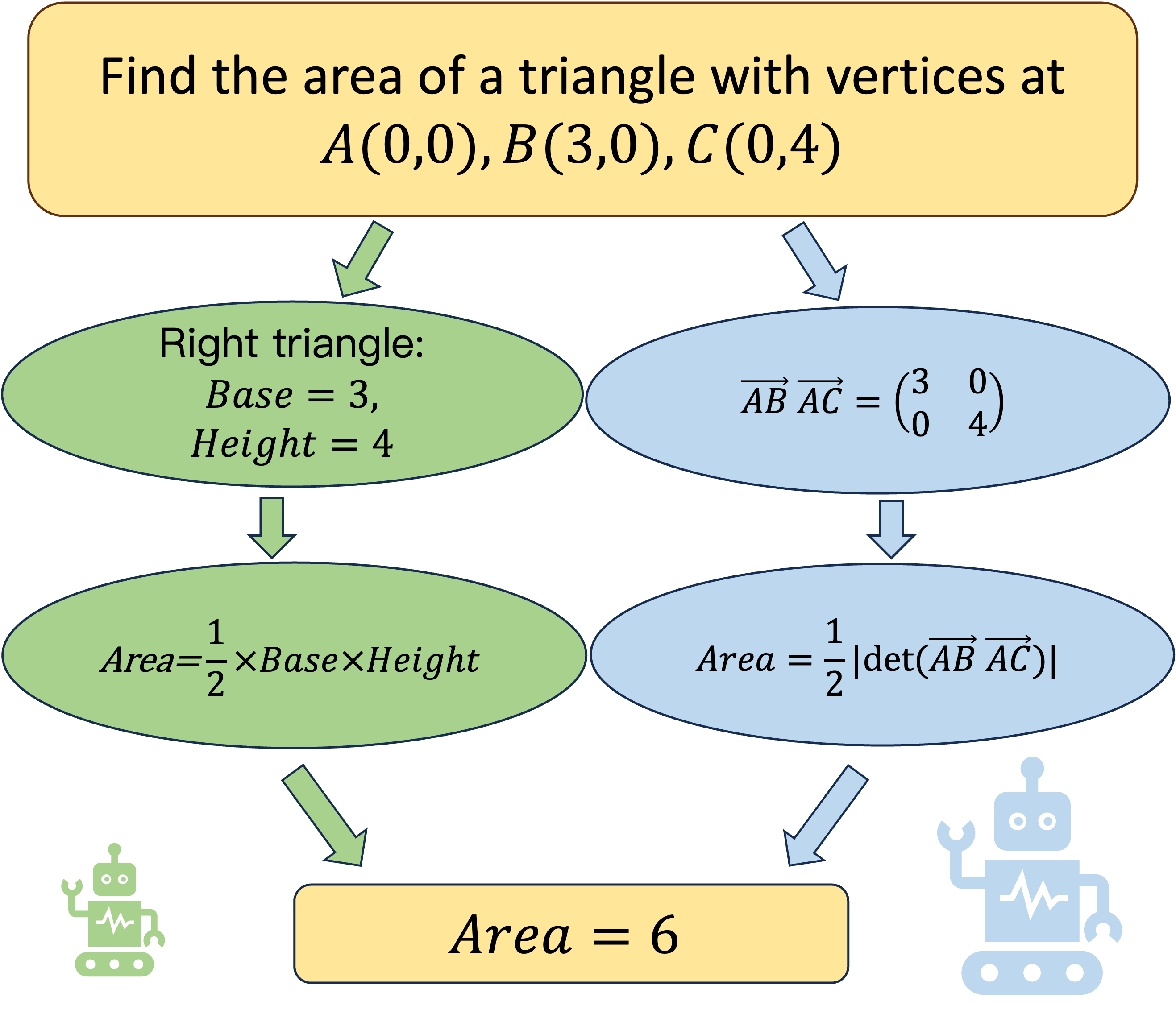}
\caption{A motivating example. Large models (right) apply vector-based algebraic abstraction to solve the problem, while small models (left) employ simple formulaic geometric decomposition. This trajectory mismatch underscores the inefficacy of direct CoT distillation across models with substantial capacity gaps.}
\label{fig:cognitive-trajectories}
\end{figure}

A straightforward approach to address this challenge is the direct distillation of Chain-of-Thought (CoT) rationales~\cite{DBLP:conf/nips/Wei0SBIXCLZ22} or other deep thoughts (such as Tree-of-Thought~\cite{DBLP:conf/nips/YaoYZS00N23}) from larger LRMs to smaller ones. This technique is widely applied to improve the capabilities of smaller LRMs~\cite{DBLP:conf/acl/HsiehLYNFRKLP23,DBLP:journals/corr/abs-2212-00193,DBLP:journals/corr/abs-2312-10730,DBLP:conf/emnlp/YueWHW24}. However, smaller models\footnote{In this work, we regard smaller LLMs as decoder-only language models typically with fewer than 10B parameters.} inherently exhibit different reasoning capacities and cognitive trajectories when solving problems compared to their larger counterparts, as illustrated in Figure~\ref{fig:cognitive-trajectories}. Similar findings have also been presented in~\cite{DBLP:journals/corr/abs-2210-06726,DBLP:conf/iclr/0006LCF24,DBLP:conf/aaai/HuHWZSN24,DBLP:conf/llm4code/LiCXJLTLL24}. This phenomenon indicates that direct distillation of CoTs from larger models can sometimes be ineffective due to the large capacity gap. Thus, a natural question arises: \emph{How can we improve the reasoning abilities of smaller LRMs in a way that is aligned with their own cognitive capacity?}

In this paper, we introduce the ``Critique-Rethink-Verify'' (CRV) system, a novel approach to enhance the reasoning capabilities of smaller models. CRV leverages multiple LLM agents, each with specialized functions working in synergy: (i) critiquing CoT rationales by considering the cognitive limits of smaller LRMs, (ii) rethinking and refining these CoTs, integrating the feedback from previous critiques, and (iii) verifying the accuracy and validity of the refined reasoning paths. Extending the Direct Preference Optimization (DPO) technique~\cite{DBLP:conf/nips/RafailovSMMEF23}, we further propose the Cognitive Preference Optimization (CogPO) algorithm to align the reasoning process with the cognitive capacities of smaller LRMs, building upon the CRV system. Ultimately, the reasoning performance of smaller models can be improved effectively.

We evaluate the effectiveness of our approach on several challenging reasoning benchmarks that are difficult for models with limited parameter sizes, such as AIME 2024, MATH-500~\cite{DBLP:journals/corr/abs-2305-20050}, GPQA-Diamond~\cite{DBLP:journals/corr/abs-2311-12022}, and LiveCodeBench. The results indicate that the small LRMs trained using the CRV+CogPO framework achieve outstanding reasoning performance.

In summary, our major contributions are:
\begin{itemize}
    \item We present the CRV system for training small yet powerful LRMs, based on multiple LLM agents specialized in unique tasks.
    \item We propose the CogPO algorithm, which continuously enhances the reasoning abilities of small models by aligning their reasoning processes with their cognitive capacities.
    \item Evaluations on challenging benchmarks demonstrate that the CRV+CogPO framework significantly improves the reasoning performance of small models, outperforming other popular training methods.
\end{itemize}

\section{Related Work}

In this section, we summarize the related work in the following three aspects.

\subsection{Prompting LLMs to Reason}
Prompting strategies to improve reasoning in large language models (LLMs) have become a critical focus. Initial studies showed that LLMs could perform basic reasoning tasks using carefully crafted prompts, such as linguistic analysis~\cite{chen-etal-2021-neurallog} and commonsense inference~\cite{latcinnik2020explaining,shwartz-etal-2020-unsupervised}. 
Chain-of-Thought (CoT)~\cite{cot} prompting explicitly guides LLMs through step-by-step reasoning, enabling them to decompose complex problems into manageable intermediate steps. Tree-of-Thought (ToT)~\cite{tot} prompting introduces a hierarchical structure to reasoning trajectories, allowing models to explore multiple solution paths. Furthermore, self-refine~\cite{Reflexion2023,Self-Refine2023} prompting incorporates verification checkpoints, where models validate intermediate results before advancing.

\begin{figure*}
\centering
\includegraphics[width=0.985\textwidth]{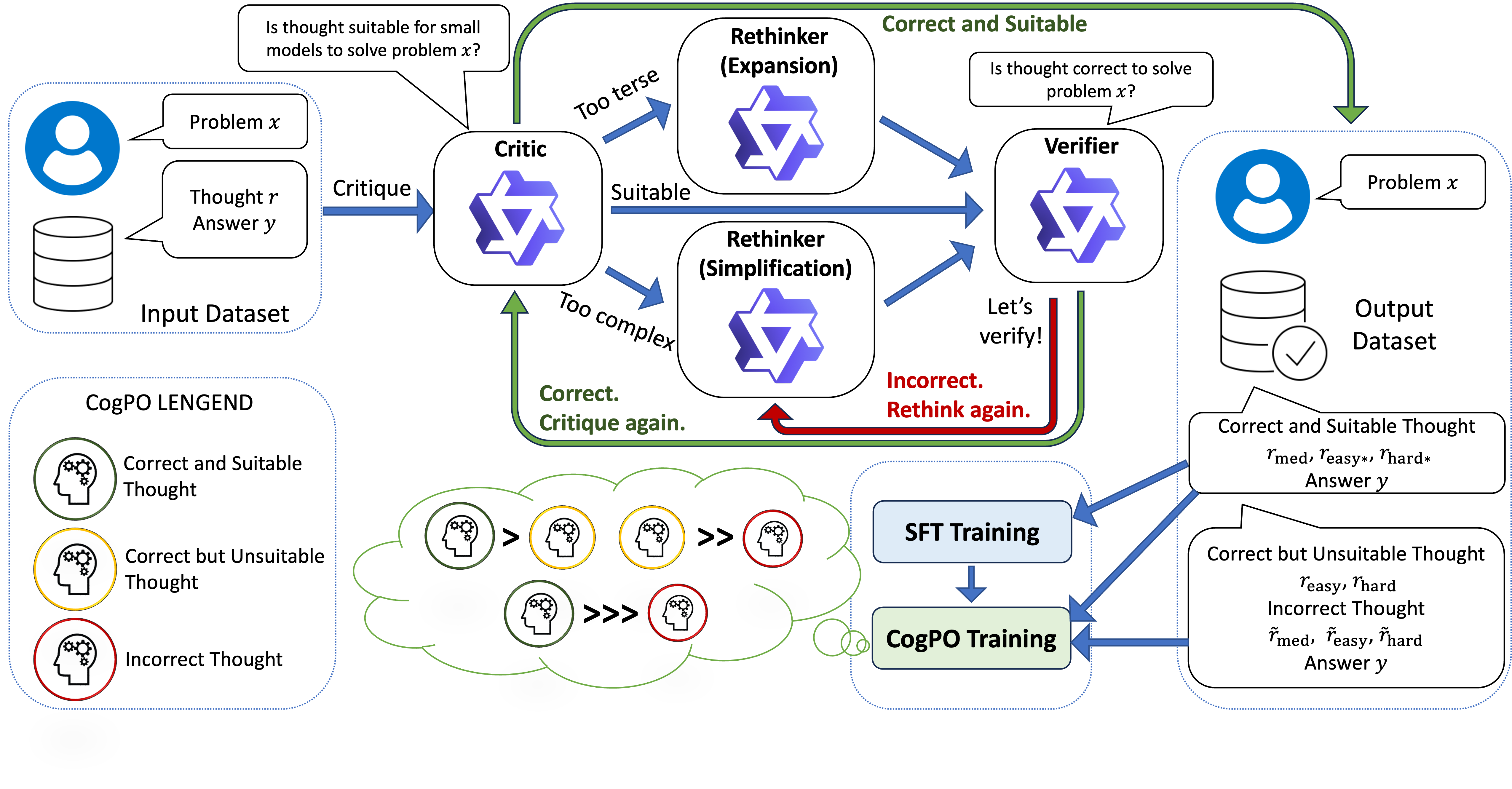}
\caption{Overview of our CRV+CogPO framework, consisting of two synergistic phases: (1) SFT training with cognitively aligned data generated by the CRV system, and (2) CogPO: dynamic $\beta$ adjustment preference optimization training using cognitive reasoning pairs with different quality gaps. 
\textbf{Disclaimer:} We use the Qwen logo as our backbone; however, any LLMs with sufficient capabilities can serve as agents as well.}
\label{fig:framework}
\end{figure*}

\subsection{Reasoning LLMs}
With the advancement of LLMs, model capabilities have steadily improved~\cite{DBLP:journals/corr/abs-2409-06857,DBLP:journals/corr/abs-2408-16737}. Models with approximately 7B to 14B parameters now show remarkable performance, and their fine-tuning costs have become increasingly feasible. This has led to the emergence of specialized small models tailored for mathematical and code-related reasoning tasks such as Qwen-Math\footnote{\url{https://qwenlm.github.io/blog/qwen2.5-math/}}, Qwen-Coder\footnote{\url{https://qwenlm.github.io/blog/qwen2.5-coder-family/}}, and Macro-o1~\cite{DBLP:journals/corr/abs-2411-14405}.

Recent studies~\cite{ShridharSS23,Yan-c2s,DBLP:journals/corr/abs-2410-05318,DBLP:journals/corr/abs-2404-02078,DBLP:journals/corr/abs-2504-09100} have investigated fine-tuning methods to enhance the reasoning abilities of smaller models. By utilizing intermediate reasoning steps, LLMs can iteratively refine their outputs~\cite{DBLP:conf/acl/JiangSYL0LK24,DBLP:conf/acl/0001FLS0XWBLJCS24,DBLP:conf/coling/ChenWZ25}. This methodology facilitates the development of small reasoning models, particularly following the release of stronger reasoning models such as DeepSeek-R1~\cite{deepseekr1}, QwQ-32B\footnote{\url{https://qwenlm.github.io/blog/qwq-32b/}} and many others. 

\subsection{Alignment Training}
To effectively train LLMs, a reinforcement learning stage is typically employed after the supervised fine-tuning (SFT) phase, serving to improve the model's alignment toward specific objectives. Reinforcement learning from human feedback (RLHF)~\cite{DBLP:conf/nips/Ouyang0JAWMZASR22} has shown effectiveness in aligning LLMs with human preferences. A potential drawback of RLHF is the explicit need for a reward model and the instability of RL training. Direct Preference Optimization (DPO)~\cite{DBLP:conf/nips/RafailovSMMEF23} trains LLMs based on selected and rejected responses. Since the introduction of DPO, several approaches have been proposed to enhance its efficacy and efficiency. For example, CPO~\cite{DBLP:conf/icml/XuSCTSDM024} extends DPO to avoid generating adequate but not perfect machine translations. SimPO~\cite{DBLP:journals/corr/abs-2405-14734} simplifies DPO by eliminating the reference model. KTO~\cite{DBLP:journals/corr/abs-2402-01306} and NCA~\cite{DBLP:journals/corr/abs-2402-05369} develop novel optimization objectives that leverage unpaired data for model alignment. Furthermore, SPPO~\cite{DBLP:journals/corr/abs-2405-00675} employs on-policy sampling to generate preference data, often outperforming off-policy DPO methods. In our work, we extend DPO to align reasoning abilities with the cognitive limits of small LLMs.

\section{Proposed Approach}

In this section, we present the techniques of our CRV system and the CogPO training algorithm.

\subsection{Overall Framework}
Our framework consists of two synergistic phases: (1) SFT with cognitively aligned data generated by the CRV system, and (2) CogPO with dynamic $\beta$ adjustment. As illustrated in Figure~\ref{fig:framework}, the CRV system first refines data tailored to the cognitive capacity of smaller LRMs for SFT training, and CogPO further aligns reasoning preferences through suitability-aware optimization using pairs with different quality gaps. This design ensures that the model initially acquires capacity-matched reasoning patterns, followed by refinement of its decision boundaries through gap-sensitive learning. Here, the decision boundary refers to the model's ability to judge whether the produced CoT is correct and aligned with its own cognitive capabilities, enabling it to successfully solve problems by following its CoT.

\subsection{The CRV System}
The CRV system employs LLM agents to construct an SFT dataset aligned with the cognitive limits of the smaller models to be trained. The input to the CRV system is an initial training set $\mathcal{D}_{\text{SFT}} = \{(x, y, r_{\text{orig}})\}$, where the three elements denote the problem, the correct answer, and the original reasoning process generated by any large LRM (e.g., DeepSeek-R1), which has been validated as correct. The following describes each agent in the CRV system.

\subsubsection{Critic}
An LLM agent first evaluates the appropriateness of reasoning processes for the target small LLM (denoted as $\pi_{\text{base}}$). For each $(x, y, r_{\text{orig}}) \in \mathcal{D}_{\text{SFT}}$, the Critic assesses $r_{\text{orig}}$ using the criterion of \textit{Cognitive Matching Degree}, checking whether the complexity and difficulty of $r_{\text{orig}}$ aligns with the cognitive capacity of $\pi_{\text{base}}$. Specifically, the Critic classifies the reasoning processes into three subsets: i) $\mathcal{D}_{\text{easy}}$ : $(x, y, r_{\text{easy}})$, cases where the reasoning process is overly terse, making it difficult for $\pi_{\text{base}}$ to follow; ii) $\mathcal{D}_{\text{med}}$ : $(x, y, r_{\text{med}})$, cases with appropriate steps that enable successful problem solving; and iii) $\mathcal{D}_{\text{hard}}$ : $(x, y, r_{\text{hard}})$, cases with overly redundant or excessively complex reasoning steps that exceed the comprehension of $\pi_{\text{base}}$, making it extremely likely to fail to guide $\pi_{\text{base}}$ in solving $x$.

\noindent\underline{\emph{Remarks.}} An intuitive approach would be to use $\pi_{\text{base}}$ itself as the Critic. However, due to its small parameter size (e.g., 7B), certain CoTs exceed $\pi_{\text{base}}$'s comprehension, rendering it incapable of reliable complexity classification. Thus, we leverage the same LLM for the Rethinker (denoted as $\pi_{\text{large}}$) to serve as the Critic, forcing it to ``think'' from the perspective of the small model $\pi_{\text{base}}$. A detailed analysis of choices for the Critic is provided in the Experiments (Section~\ref{experiment:Choices of the Critic}) and Appendix~\ref{sec:criticAnalysis}. 

\noindent\underline{\emph{Hypothesis Verification.}} 
To further verify that the complexity levels of CoTs are closely related to the cognitive capacities of reasoning models, we conduct an experiment in which we evaluate DeepSeek-R1-Distill-Qwen-1.5B/7B/32B on MATH500, collecting each model’s outputs. We employ the Critic to rate the level of each model’s CoT outputs; each CoT is evaluated three times, and the final rating is determined by majority vote. For each model, we quantify the distribution of these CoTs across different complexity levels in Table~\ref{tab:percent}. As shown, DeepSeek-R1-Distill-Qwen-1.5B yields the largest number of simple CoTs, while DeepSeek-R1-Distill-Qwen-32B generates the greatest number of difficult CoTs. 

These findings demonstrate that the complexity of CoTs escalates as the model size increases, suggesting that larger models possess higher reasoning and cognitive capacities. Consequently, overly terse or complex CoTs may not be suitable for training models with lower cognitive abilities. It is therefore essential to use CoTs that align with the model's cognitive trajectory to improve its reasoning capabilities, a strategy akin to ``teaching according to the student's ability.''

\begin{table}[t]
\centering
\begin{tabular}{l | lll}
\toprule
\bf Level/Model Size & \bf 1.5B & \bf 7B & \bf 32B\\ 
\midrule
Easy & 195 & 80 & 19\\
Medium & 296 & 389 & 354\\
Hard & 9 & 31 & 127\\
\bottomrule
\end{tabular}
\caption{Complexity distributions of CoTs generated by different sizes of DeepSeek-R1-Distill-Qwen models.}
\label{tab:percent}
\end{table}

\subsubsection{Rethinker}
An LLM agent $\pi_{\text{large}}$ is tasked with rewriting reasoning processes to achieve cognitive alignment. For each $(x, y, r_{\text{easy}}) \in \mathcal{D}_{\text{easy}}$, the Rethinker expands $r_{\text{easy}}$ by adding necessary steps for easier understanding, i.e., $r_{\text{easy*}} = \pi_{\text{large}}(x, y, r_{\text{easy}})$. Similarly, for each $(x, y, r_{\text{hard}}) \in \mathcal{D}_{\text{hard}}$, the Rethinker simplifies $r_{\text{hard}}$ by removing redundancies or using simpler methods to solve the problem grounded in the correct answer: $r_{\text{hard*}} = \pi_{\text{large}}(x, y, r_{\text{hard}})$. Examples of the rewriting process of the Rethinker are shown in Tables~\ref{tab:case1} and~\ref{tab:case2}.

\subsubsection{Verifier}
Finally, we leverage the LLM agent $\pi_{\text{base}}$ to validate the correctness of $r_{\text{med}}$, $r_{\text{easy*}}$, and $r_{\text{hard*}}$ in order to preserve the high quality of the dataset. It predicts whether $\pi_{\text{base}}$ can derive the correct answer $y$ from the rewritten thoughts $r_{\text{easy*}}$ or $r_{\text{hard*}}$. Note that $r_{\text{med}}$ has already been validated as correct in the original dataset, and we send $r_{\text{med}}$ to the Verifier to further ensure data quality.

After verification, incorrect cases are sent back to the Rethinker to be continuously rewritten until they pass verification. In the implementation, cases that fail after three iterations are discarded. For cases that pass verification, we invoke the Critic to make the judgment again (please refer to Figure~\ref{fig:framework} for the algorithmic flow). 

The final SFT dataset is composed of verified medium-level data: $\mathcal{D}_{\text{SFT*}} = \mathcal{D}_{\text{med}} \cup \mathcal{D}_{\text{easy*}} \cup \mathcal{D}_{\text{hard*}}$, where $\mathcal{D}_{\text{med}}$ denotes the verified medium-level data, and $\mathcal{D}_{\text{easy*}}$ and $\mathcal{D}_{\text{hard*}}$ represent the rewritten versions of $\mathcal{D}_{\text{easy}}$ and $\mathcal{D}_{\text{hard}}$ that have passed verification and have been re-rated as medium by the Critic, respectively. $\mathcal{D}_{\text{SFT*}}$ serves as the SFT training set in the CRV stage. Prompt templates used in the CRV system are provided in Appendix~\ref{sec:prompts}.

\subsection{Cognitive Preference Optimization}

The CogPO algorithm aligns CoT processes of smaller LLMs with their inherent cognitive capacities, following the SFT training using the CRV system.

\subsubsection{Preliminaries}
Briefly speaking, the CogPO algorithm is extended from DPO~\cite{DBLP:conf/nips/RafailovSMMEF23} and its variants. Let $y_w$ and $y_l$ be the chosen and rejected responses for an instruction $x$ (not restricted to reasoning problems addressed in this work), respectively. We further denote $\pi_{\theta}$ as the model to be optimized after SFT and $\pi_\text{ref}$ as the reference model. DPO seeks to maximize the following margin:

\begin{small}
\begin{equation}
M_{\beta}(x, y_w, y_l) = \beta \cdot \left( \log \frac{\pi_{\theta}(y_w|x)}{\pi_\text{ref}(y_w|x)} - \log \frac{\pi_{\theta}(y_l|x)}{\pi_\text{ref}(y_l|x)} \right)
\end{equation}
\end{small}
where $\beta$ is a temperature hyperparameter. Based on $M_{\beta}(x, y_w, y_l)$, the DPO loss is defined as:
\begin{equation}
    \mathcal{L}_\text{DPO} = -\mathbb{E}_{(x, y_w, y_l) \sim \mathcal{D}} \log \sigma(M_{\beta}(x, y_w, y_l)).
\end{equation}
The setting of $\beta$ is critical to the performance of DPO. $\beta$-DPO~\cite{DBLP:journals/corr/abs-2407-08639} further adjusts $\beta$ according to $M_{\beta}(x, y_w, y_l)$, either at the instance level or batch level, allowing the model to adapt $\beta$ based on the reward differential of the input data.

\subsubsection{Algorithmic Description}
As noted, DPO and $\beta$-DPO do not require any prior knowledge of how the model learns the user's preferences. We suggest that this type of prior knowledge is critical for training better small reasoning models, as the cognitive trajectories of large and small models often differ~\cite{DBLP:journals/corr/abs-2210-06726,DBLP:conf/iclr/0006LCF24,DBLP:conf/aaai/HuHWZSN24}, and this may not be directly reflected in the reward differential. We propose CogPO to align reasoning preferences by encoding more prior knowledge and continuously training on a series of \emph{mini-tasks}.

\begin{figure}
\centering
\includegraphics[width=.9\columnwidth]{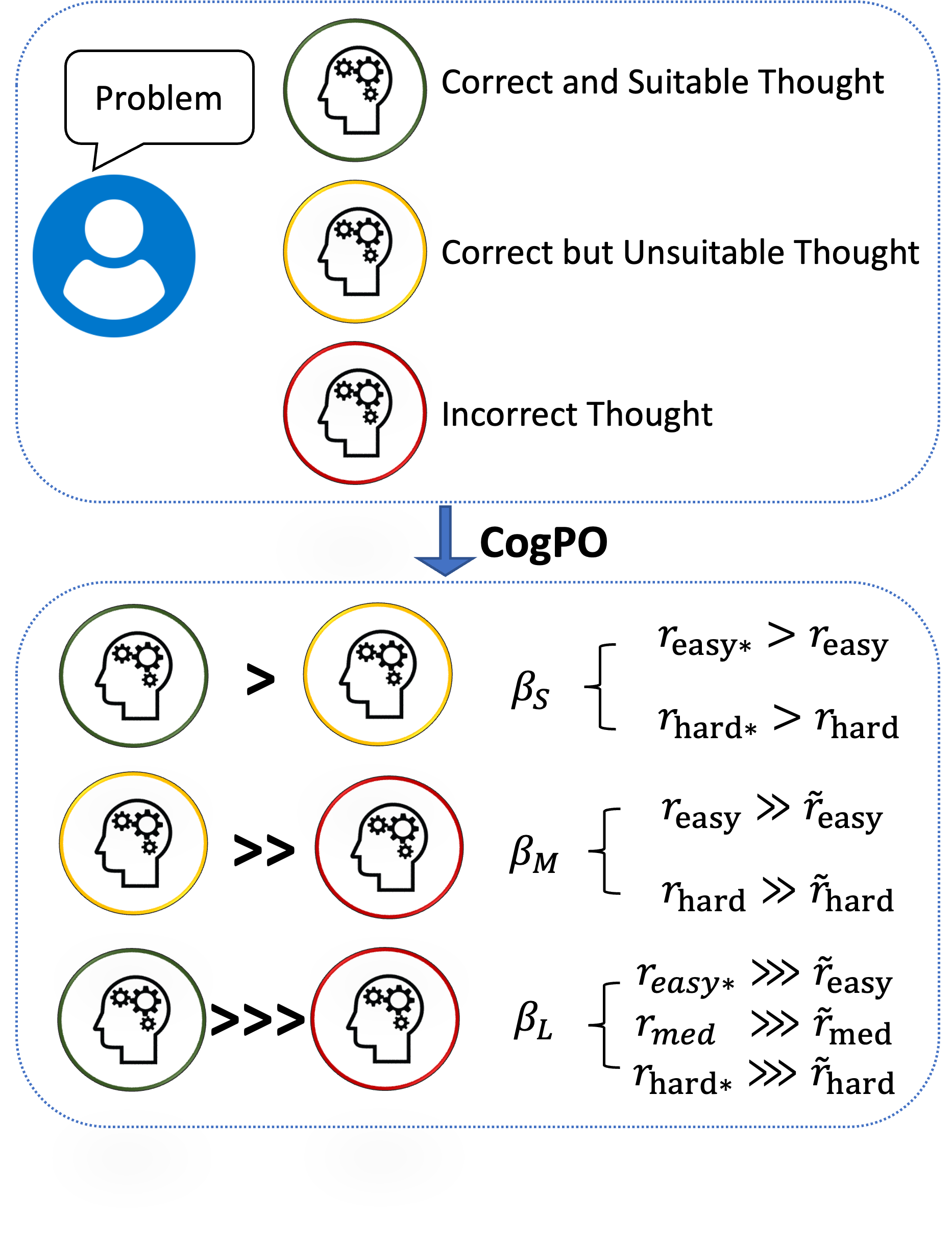}
\caption{An illustration of the proposed CogPO algorithm, showing the different preference gaps between CoT pairs and the corresponding mini-tasks.}
\label{fig:cogpo}
\end{figure}

\begin{table*}
\centering
\begin{small}
\begin{tabular}{l | l | ll | llll}
\toprule
\bf Dataset/Model & \bf Zero-shot & \bf SFT & \bf CRV+SFT  & \bf DPO  & \bf $\beta$-DPO & \bf SimPO &\bf CogPO\\ 
\midrule
AIME2024 & 10.0 & 20.0 & \bf 26.7 & 23.3 & 23.3 & \underline{26.7} & \bf 30.0\\
MATH-500 & 73.6	& 80.0 & \bf 84.0 & 83.4 & 83.8 & \underline{84.2} & \bf 84.4\\
GSM8K & 89.5 & 92.3 & \bf 92.7 & 92.6 & \underline{93.0} & 92.6 & \bf 93.3\\
GPQA Diamond & 33.3 & 37.4 & \bf40.9 & 40.0 & 37.4 & \bf 40.9 & \bf 40.9\\
LiveCodeBench V2 & 30.7	& 31.3 & \bf 34.4 & 34.4 & 35.8 & \underline{36.2} & \bf 36.6\\
MMLU & 71.9 & 76.1 & \bf 76.5& 76.1 & 76.0 & \bf 76.5 & \bf 76.5\\
OlympiadBench (math-en) & 40.1 & 43.6 & \bf 45.8 & 45.7 & \underline{46.5} & 46.0 & \bf 46.6\\
\bottomrule
\end{tabular}
\end{small}
\caption{Performance comparison of various training methods. The LLM backbone is Qwen2.5-7B-Instruct, and the training set is Bespoke-Stratos-17k. Results are shown for zero-shot (without further training), SFT, CRV+SFT, DPO, $\beta$-DPO, SimPO, and CogPO. DPO, $\beta$-DPO, SimPO, and CogPO are conducted on the same model checkpoints of CRV+SFT, using the same preference pair dataset. The metrics represent scores for these tasks, with the best results for each dataset in each group marked in bold and the second-best underlined.}
\label{tab:main}
\end{table*}

\begin{table*}
\centering
\begin{small}
\begin{tabular}{l|ll | ll|ll}
\toprule
\bf Dataset/Model & \bf LLaMA-O1 & \bf Macro-o1 & \bf Bespoke-Stratos-7B & \bf Ours & \bf OpenThinker-7B & \bf Ours \\ 
\midrule
Training Set Size & 332K & 60K & 17K &  17K & 114K  & 114K\\
\midrule
AIME2024 &  3.3 &  6.7 & 20.0 & \bf 30.0 &  31.3 & \bf 43.3\\
MATH500 &  28.6 &  38.4 & 82.0 & \bf 84.4 & 83.0 & \bf 88.4\\
GPQA Diamond &  26.3 &  31.8 & 37.8 & \bf 40.9 & 42.4 & \bf 42.9\\
LiveCodeBench V2 & 1.6 & 24.9 & 36.1 & \bf 36.6 &  39.9 & \bf 46.4\\
\bottomrule
\end{tabular}
\end{small}
\caption{Comparison between our model and other small reasoning models in the open-source community. Specifically, we train two versions using our approach on Bespoke-Stratos-17k and OpenThoughts-114k, respectively, where the two training sets are the same as Bespoke-Stratos-7B and OpenThinker-7B, respectively.}
\label{tab:other}
\end{table*}

We leverage the Rethinker in CRV to also generate incorrect reasoning processes when prompted to rewrite the original thought $r_{\text{orig}}$ (prompt template is provided in Appendix~\ref{sec:prompts}). The incorrect thoughts are denoted as $\tilde{r}_{\text{med}}$, $\tilde{r}_{\text{easy}}$, and $\tilde{r}_{\text{hard}}$, based on their origin from $\mathcal{D}_{\text{med}}$, $\mathcal{D}_{\text{easy}}$, and $\mathcal{D}_{\text{hard}}$. These thoughts contain factual errors or invalid reasoning steps, which can mislead $\pi_{\text{base}}$, rendering it impossible to solve $x$. Thus, we categorize the properties of all the thoughts collected into the following three types:
i) $r_{\text{med}}$, $r_{\text{easy*}}$, and $r_{\text{hard*}}$: medium-level reasoning processes that are both correct and cognitively suitable for $\pi_{\text{base}}$; 
ii) $r_{\text{easy}}$ and $r_{\text{hard}}$: easy or hard thoughts that are correct but unsuitable for $\pi_{\text{base}}$; 
iii) $\tilde{r}_{\text{med}}$, $\tilde{r}_{\text{easy}}$, and $\tilde{r}_{\text{hard}}$: incorrect reasoning processes with logical flaws or invalid reasoning steps (regardless of the difficulty levels). To define the mini-tasks used for CogPO training, we consider the preference gaps in these three types of CoT pairs as follows:

\begin{itemize}
    \item \textbf{Small Gap Mini-task}: The pairs are ($r_{\text{easy*}}$, $r_{\text{easy}}$) and ($r_{\text{hard*}}$, $r_{\text{hard}}$). Both are correct but differ in complexity (suitable vs. unsuitable for $\pi_{\text{base}}$). We treat $r_{\text{easy*}}$ and $r_{\text{hard*}}$ as chosen reasoning processes ($r_w$), and $r_{\text{easy}}$ and $r_{\text{hard}}$ as rejected ($r_l$).

    \item \textbf{Medium Gap Mini-task}: The pairs are ($r_{\text{easy}}$, $\tilde{r}_{\text{easy}}$) and ($r_{\text{hard}}$, $\tilde{r}_{\text{hard}}$). The former are correct but unsuitable, while the latter are completely incorrect. As correctness is more important than suitability for our model, the preference gap of this mini-task should be higher than that in the previous case. For this mini-task, $r_{\text{easy}}$ and $r_{\text{hard}}$ are treated as $r_w$, while $\tilde{r}_{\text{easy}}$ and $\tilde{r}_{\text{hard}}$ are treated as $r_l$.

    \item \textbf{Large Gap Mini-task}: The pairs are ($r_{\text{med}}$, $\tilde{r}_{\text{med}}$), ($r_{\text{easy*}}$, $\tilde{r}_{\text{easy}}$), and ($r_{\text{hard*}}$, $\tilde{r}_{\text{hard}}$). Intuitively, the preference gaps should be the largest between suitable and correct thoughts and incorrect ones. Here, $r_{\text{med}}$, $r_{\text{easy*}}$, and $r_{\text{hard*}}$ are treated as $r_w$, while $\tilde{r}_{\text{med}}$, $\tilde{r}_{\text{easy}}$, and $\tilde{r}_{\text{hard}}$ are treated as $r_l$.
\end{itemize}

Following our modeling framework, each training instance $(x, r_w, r_l)$ receives its specific $\beta$ value, as illustrated in Figure~\ref{fig:cogpo}. The CogPO objective function aggregates these preferences:
\begin{equation}
\mathcal{L}_\text{CogPO} = -\mathbb{E}_{(x, r_w, r_l) \sim \mathcal{D}} \log \sigma(M_{\beta_{\text{CogPO}}}(x, r_w, r_l)),
\end{equation}
where $\beta_{\text{CogPO}}$ is selected from $\{\beta_{\text{S}}, \beta_{\text{M}}, \beta_{\text{L}}\}$, depending on the specific types of mini-tasks (with $\beta_{\text{S}} < \beta_{\text{M}} < \beta_{\text{L}}$, corresponding to the three gaps).

Overall, our CogPO algorithm enables granular preference learning: strong regularization ($\beta_{\text{L}}$) for validity discrimination, moderate guidance ($\beta_{\text{M}}$) for suitability alignment, and subtle refinement ($\beta_{\text{S}}$) for reasoning style adaptation. This design provides more control over the alignment process, leading to further improvements on the basis of SFT (using the CRV system).

\noindent\underline{\emph{Remarks.}} CogPO can be naturally combined with $\beta$-DPO~\cite{DBLP:journals/corr/abs-2407-08639}. We can redefine the $\beta$ values $\{\beta_{\text{S}}, \beta_{\text{M}}, \beta_{\text{L}}\}$ as follows:
\begin{equation}
\beta_i^* = \beta_i + \alpha \cdot (M_i - M_0) \cdot \beta_i
\end{equation}
where $\beta_i$ is chosen from $\{\beta_{\text{S}}, \beta_{\text{M}}, \beta_{\text{L}}\}$ based on the corresponding gap type, $M_i$ is the instance-level reward differential, and $M_0$ is a predefined threshold as in~\citet{DBLP:journals/corr/abs-2407-08639}.\footnote{In our experiments, this combination does not yield substantial improvements, as prior knowledge is more important for our task. Hence, we stick to using $\mathcal{L}_\text{CogPO}$.}

\section{Experiments}

To evaluate the effectiveness of the CRV framework and the CogPO algorithm, we conduct a series of experiments on several challenging reasoning benchmarks.
Due to space limitations, the datasets and experimental settings are described in Appendix~\ref{appendix: Datasets} and~\ref{appendix: Experimental Details}.

\begin{table*}
\centering
\begin{small}
\begin{tabular}{l | llllll}
\toprule
\bf Model Backbone (The Critic) & \bf AIME2024 & \bf MATH-500 & \bf GPQA-D & \bf GSM8K & \bf LCB V2 & \bf OlympiadBench\\ 
\midrule
Qwen2.5-7B-Instruct & 13.3 & 80.2 & \bf 40.9 & 92.3 & 30.5 & 43.9\\
Qwen2.5-32B-Instruct & 23.3 & 82.2 & 39.9 & 92.6 & 33.3 & 45.1\\
Qwen2.5-72B-Instruct & 20.0 & 81.8 & 36.4 & \bf 92.7 & 30.5 & 42.0\\
DeepSeek-R1-Distill-Qwen-32B & \bf 26.7 & \bf 84.0 & \bf 40.9 & \bf 92.7 & \bf 34.4 & \bf 45.8\\
\bottomrule
\end{tabular}
\end{small}
\caption{Comparison using different backbones as the Critic. All results are produced using CRV+SFT without CogPO on Bespoke-Stratos-17k.}
\label{tab:critic}
\end{table*}

\subsection{Main Experimental Results and Ablations}

We choose Bespoke-Stratos-17k as our training set. Table~\ref{tab:main} presents the results of our CRV framework and the CogPO algorithm on various reasoning benchmarks. CRV+SFT outperforms direct SFT on all benchmarks. Building on CRV+SFT, CogPO further enhances the model’s reasoning ability, surpasses other preference‐optimization algorithms, and ultimately achieves the strongest performance, demonstrating its effectiveness in aligning the model's reasoning processes with its cognitive capacities. These results reveal that our CRV+CogPO framework effectively enhances the reasoning capabilities of smaller models, outperforming other methods by a large margin.

\subsection{Comparison Against Other Models}

We compare our trained 7B model with other models released in the open-source community. We consider two reasoning LLMs available before the launch of DeepSeek-R1, namely Macro-o1~\cite{DBLP:journals/corr/abs-2411-14405} and LLaMA-O1\footnote{\url{https://huggingface.co/SimpleBerry/LLaMA-O1-Supervised-1129}}. We also compare with models trained on datasets distilled from DeepSeek-R1, including Bespoke-Stratos-7B\footnote{\url{https://huggingface.co/bespokelabs/Bespoke-Stratos-7B}} and OpenThinker-7B\footnote{\url{https://huggingface.co/open-thoughts/OpenThinker-7B}}.
Using our CRV+CogPO framework, we additionally train two models on the Bespoke-Stratos-17k and OpenThoughts-114k training sets, respectively. Thus, it is fair to compare our method against Bespoke-Stratos-7B and OpenThinker-7B. The results, along with the sizes of the training sets, are shown in Table~\ref{tab:other}. It can be observed that employing DeepSeek-R1-generated CoT data yields superior results.
At the algorithmic level, both Bespoke-Stratos-7B and our model are trained on the 17K CoTs from DeepSeek-R1. Under identical data conditions, our model significantly outperforms Bespoke-Stratos-7B across all benchmarks and achieves performance comparable to OpenThinker-7B, which is trained on 114K CoTs from DeepSeek-R1. Moreover, when trained on the same dataset as OpenThinker-7B, our model substantially surpasses OpenThinker-7B on all benchmarks. These findings demonstrate that, given the same data, our CRV+CogPO training framework exhibits superior performance, confirming its effectiveness.

\begin{table}
\centering
\begin{tabular}{l | lll}
\toprule
\bf Dataset/Model & \bf Easy & \bf Medium & \bf Hard\\ 
\midrule
AIME2024 & 13.3 & 23.3 & 16.7\\
MATH500 & 75.4 & 82.8 & 78.2 \\
GPQA-D & 34.3 & 37.4 & 33.3 \\
LCB V2 & 31.9 & 36.2 & 32.5\\
\bottomrule
\end{tabular}
\caption{Experimental results on training data of different complexity levels.}
\label{tab:easy-medium-hard}
\end{table}

\subsection{Study on Choices of the Critic}
\label{experiment:Choices of the Critic}

In the previous section, we claimed that using the small target LLM $\pi_{\text{base}}$ as the Critic does not necessarily produce satisfactory results due to its limited parameter size. In contrast, larger LLMs $\pi_{\text{large}}$ can ``think like small models'' better. The results of using different backbones as the Critic are shown in Table~\ref{tab:critic}, with the backbones for the Rethinker and the Verifier unchanged. From the results, we see that they confirm our findings, as larger models consistently perform better than the 7B model in almost all tasks. Among the three large agents, DeepSeek-R1-Distill-Qwen-32B exhibits the best performance based on majority voting across all testing sets. A detailed and in-depth analysis of the selection of the Critic is provided in Appendix~\ref{sec:criticAnalysis}.

\subsection{Training with CoT Datasets of Different Complexity Levels}
To further investigate whether medium-level data are indeed the most suitable for the base model, we conduct experiments on the OpenThoughts-114K dataset. We use the Critic to rate all CoTs in the dataset, then randomly sampled 10K CoTs from each of the derived easy, medium, and hard subsets to construct three training sets. We then perform SFT with Qwen2.5-7B-Instruct on these three training sets under identical configurations. The results are shown in Table~\ref{tab:easy-medium-hard}, indicating that when the number of training data is the same, the model trained on the medium subset achieves the highest scores, fully supporting our hypothesis. The CoTs in the easy and hard sets are either too terse or overly complex, preventing the base model from effectively comprehending all CoTs in those sets. In contrast, the medium-level subset aligns with the model’s cognitive capabilities and thus yields the best results.

\begin{table}
\centering
\begin{small}
\begin{tabular}{l | llll}
\toprule
\bf Dataset/Model & \bf SFT & \bf w. C  & \bf w. CR  & \bf w. CRV\\ 
\midrule
AIME2024 & 20.0 & 23.3 & 26.7 & 26.7\\
MATH500 & 80.0 & 83.4 & 83.2 & 84.0\\
GPQA-D & 37.4 & 38.4 & 39.9 & 40.9\\
LCB V2 & 31.3 & 34.3 & 34.1 & 34.4\\
\bottomrule
\end{tabular}
\end{small}
\caption{Ablation results on the CRV system.}
\label{tab:CRV}
\end{table}

\subsection{Study on Effectiveness of Critic, Rethinker, and Verifier}

To further explore the collaborative mechanism within the CRV system and the individual roles and contributions of each component, we conduct extensive ablation experiments on the Bespoke-Stratos-17k dataset. Table~\ref{tab:CRV} presents our ablation results. The ``SFT'' row reports results from directly performing SFT on the original dataset without any CRV intervention; the ``w. C'' row shows performance when only the Critic is applied before SFT, using only the traces rated as medium by the Critic for SFT; the ``w. CR'' row indicates results when both the Critic and the Rethinker participate prior to SFT, utilizing the medium-rated traces and the refined easy/hard traces that have not yet been verified; the ``w. CRV'' row reflects outcomes when the Critic, Rethinker, and Verifier are all applied.

As the Critic, Rethinker, and Verifier are added sequentially, the model’s reasoning ability exhibits a progressively improving trend, which clearly illustrates the role of each component. Notably, ``w. CR'' experiences a performance drop on MATH500 and LCB V2, indicating that omitting the Verifier after the Rethinker’s refinement can impair the model’s reasoning ability. Therefore, each component of the CRV system plays an indispensable role. To achieve optimal performance, we recommend processing the data using the complete system.

\subsection{Study on Model Scales}

\begin{figure}
    \centering
    \begin{subfigure}{0.235\textwidth}
        \centering
        \includegraphics[width=\textwidth]{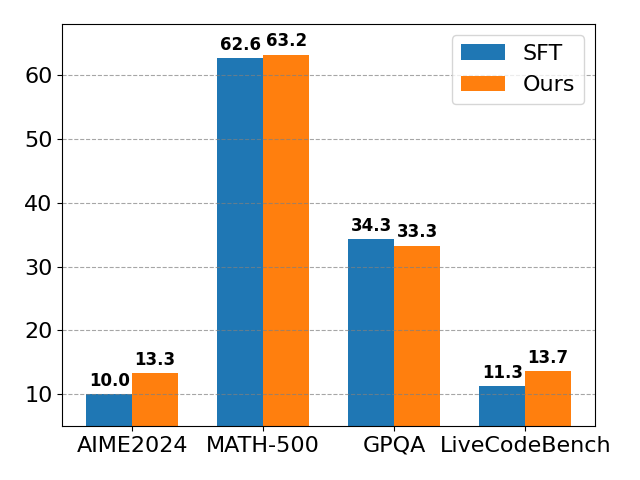}
        \caption{Qwen2.5-3B-Instruct}
    \end{subfigure}
    \hfill
    \begin{subfigure}{0.235\textwidth}
        \centering
        \includegraphics[width=\textwidth]{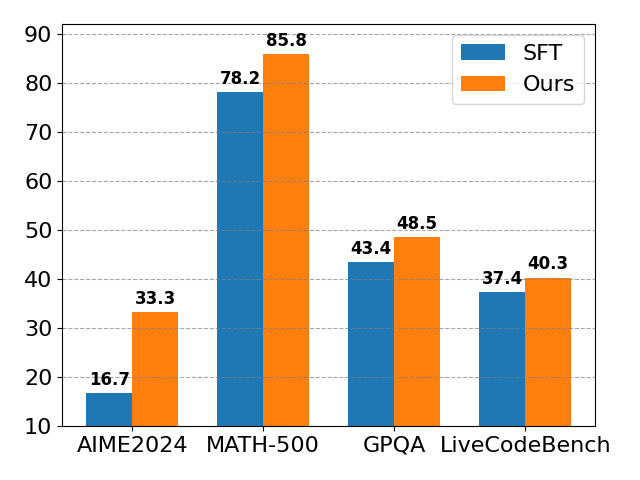}
        \caption{Qwen2.5-14B-Instruct}
    \end{subfigure}
    \caption{Experimental results for different sizes of Qwen2.5 models on AIME2024, MATH500, GPQA Diamond, and LiveCodeBench V2.}
    \label{fig:model_size}
\end{figure}

To study the effectiveness of different parameter sizes on student models, we further report the performance of Qwen2.5-3B-Instruct and Qwen2.5-14B-Instruct. The experimental settings are identical to those of Qwen2.5-7B-Instruct. The results are presented in Figure~\ref{fig:model_size}. We observe that our method is also effective across different model scales. An interesting observation is that the improvement is more significant for Qwen2.5-14B-Instruct compared to Qwen2.5-3B-Instruct. This is because, even when we leverage the CRV system to rewrite the CoTs, the large capacity gap between the teacher and student models makes it more challenging for Qwen2.5-3B-Instruct to capture the CoTs through SFT. This finding is also consistent with the recently discovered ``distillation scaling law''~\cite{distillationscalinglaws}.

\subsection{Study on Other Model Backbones}

\begin{figure}
    \centering
    \begin{subfigure}{0.235\textwidth}
        \centering
        \includegraphics[width=\textwidth]{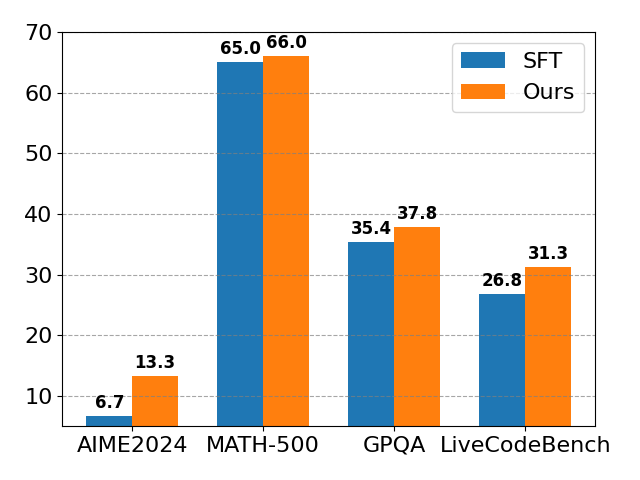}
        \caption{Llama3.1-8B-Instruct}
    \end{subfigure}
    \hfill
    \begin{subfigure}{0.235\textwidth}
        \centering
        \includegraphics[width=\textwidth]{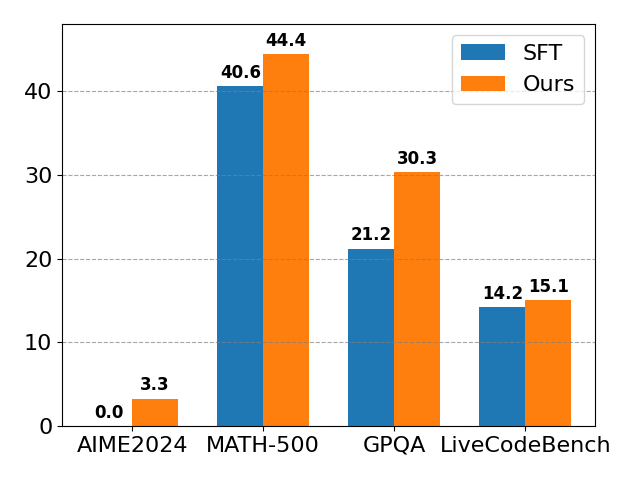}
        \caption{Mistral-7B-V0.3}
    \end{subfigure}
    \caption{Experimental results for other model series (Llama3.1-8B-Instruct, Mistral-7B-V0.3) beyond Qwen2.5, on AIME2024, MATH500, GPQA Diamond, and LiveCodeBench V2.}
    \label{fig:model_series}
\end{figure}

To evaluate the generality of the proposed approach, we perform additional experiments on multiple backbones beyond the Qwen2.5 series using the Bespoke-Stratos-17k dataset. Figure~\ref{fig:model_series} demonstrates that, for both the LLaMA and Mistral series, our approach achieves notable performance gains over the direct SFT baseline across diverse mathematical and coding tasks. These results indicate that the CRV+CogPO framework enables seamless adaptation to other backbones, demonstrating the universality of our approach and its potential to produce stronger models based on other LLMs.

\begin{figure}
    \centering
    \begin{subfigure}{0.235\textwidth}
        \centering
        \includegraphics[width=\textwidth]{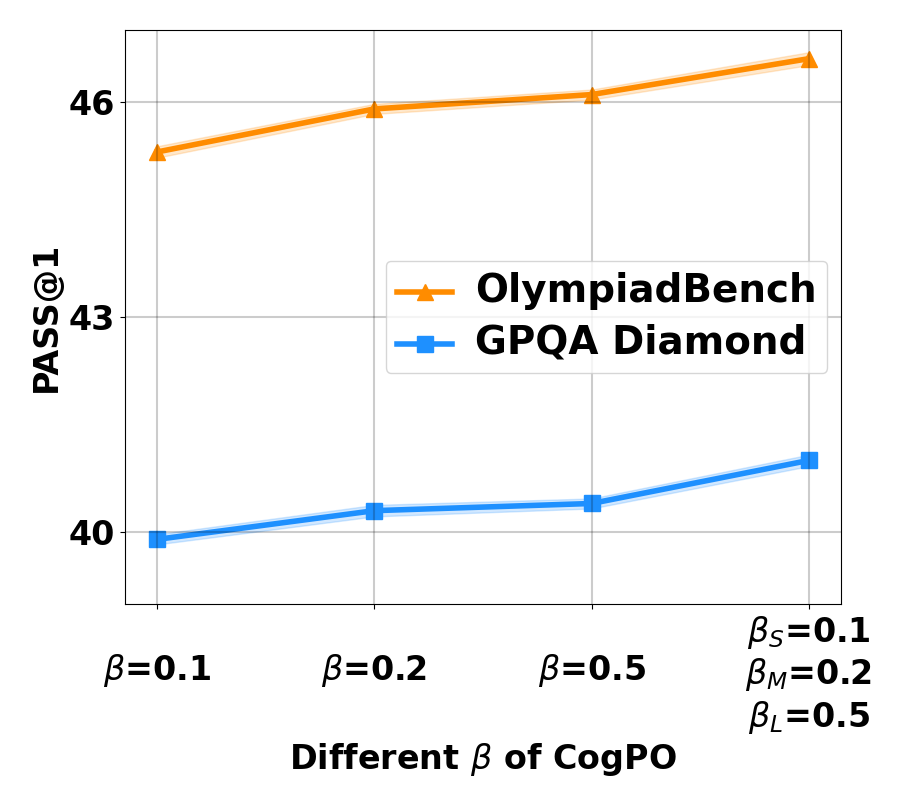}
    \end{subfigure}
    \hfill
    \begin{subfigure}{0.235\textwidth}
        \centering
        \includegraphics[width=\textwidth]{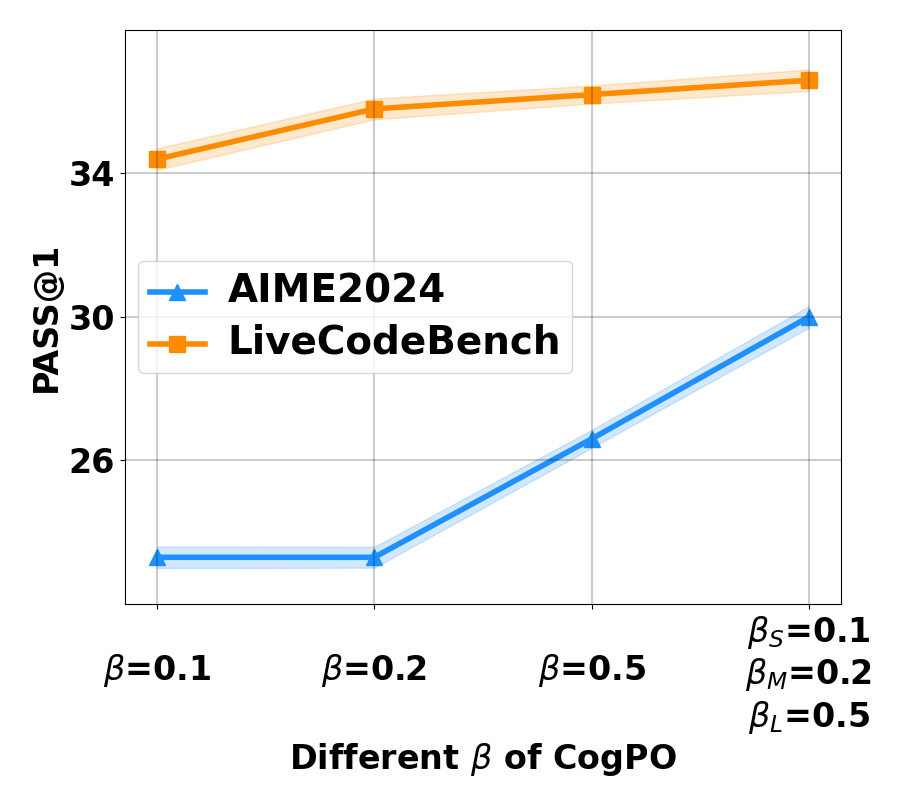}
    \end{subfigure}
    \caption{The study on impact of different $\beta$ values on AIME2024, GPQA Diamond, LiveCodeBench V2, and OlympiadBench.}
    \label{fig:dpo_beta}
\end{figure}

\subsection{Study on Computation Cost}

Our CRV+CogPO framework consists of two main stages: data processing (the CRV system) and model training (SFT and CogPO). For model training, a detailed breakdown of the computational cost is provided in Appendix~A.2. For data processing, the computational cost is primarily determined by use of the CRV system, which we explore in two key sections. Section~4.3 shows that models with stronger reasoning abilities yield superior results, while Section~4.5 confirms that every component of the CRV system positively contributes to the model's reasoning performance.

Therefore, when computational resources are ample, we recommend selecting a powerful reasoning model as the agent and utilizing the complete CRV system to construct a high-quality training dataset for optimal performance. However, if computational resources are limited, our framework can still be effectively used to enhance reasoning capabilities. The experiments in Sections~4.3 and~4.5 show that even when employing a weaker model as the Critic or using only a partial CRV system, the resulting model's reasoning ability still outperforms the original baseline.

\subsection{Hyper-parameter Analysis}

To evaluate the impact of $\beta$ values in CogPO, we perform a series of experiments with varying $\beta$ values to assess the algorithm's effectiveness. As shown in Figure~\ref{fig:dpo_beta}, the highest performance is attained when assigning tailored $\beta$ values to samples based on their respective gaps, which is a core principle of the CogPO algorithm.

\subsection{Case Studies}

Due to space limitations, case studies are included in the Appendix. These studies clearly show how our approach can effectively expand or simplify the reasoning processes based on the Critic's feedback.

\section{Conclusion and Future Work}

In this paper, we present the CRV framework where we leverage the strengths of LLM agents to critique, refine, and verify CoT outputs for optimizing CoT training sets. The CogPO algorithm further aligns model outputs with their inherent cognitive capacities, improving performance on several challenging reasoning tasks. In the future, we will (i) train and release stronger small models using larger CoT datasets, (ii) improve the effectiveness of the CRV framework, especially for much smaller models, and (iii) investigate our approach for other domain-specific applications, such as medical diagnosis and legal reasoning.

\section*{Limitations}

While our proposed framework shows promising results in enhancing the reasoning capabilities of smaller LLMs, several limitations remain. The CRV framework relies heavily on the contributions of larger models in refining the CoT outputs. This dependency may create challenges in situations where access to larger models is restricted, or when these larger models generate incorrect results. In addition, although our framework is designed for smaller LLMs, there remains a ceiling on their performance. By nature, smaller models inherently have reduced capacity to encode complex information and handle nuanced reasoning tasks, which may limit their effectiveness in certain scenarios.

\section*{Ethical Considerations}
Our work is fully methodological; hence, there are no direct ethical issues. However, smaller models trained on data distilled from larger ones might inherit or exacerbate biased outputs, which can influence outcomes. We suggest that continuous evaluation of trained LLMs based on ethical guidelines is indispensable.

\section*{Acknowledgments}

This work was supported by Alibaba Research Intern Program.

\appendix

\newpage

\section{Supplementary Experiments}

\subsection{Datasets}
\label{appendix: Datasets}

In our experiments, we evaluate our work on several benchmarks, including AIME2024\footnote{\url{https://huggingface.co/datasets/Maxwell-Jia/AIME_2024}}, MATH-500~\cite{DBLP:journals/corr/abs-2305-20050}, GSM8K~\cite{gsm8k}, GPQA Diamond~\cite{DBLP:journals/corr/abs-2311-12022}, LiveCodeBench V2~\cite{DBLP:journals/corr/abs-2403-07974}, MMLU~\cite{DBLP:conf/iclr/HendrycksBBZMSS21}, and OlympiadBench (math-en)~\cite{DBLP:conf/acl/HeLBHTSHHHZLQL024}. The sizes of our testing sets are summarized in Table~\ref{tab:stat}.

For our training set $\mathcal{D}_{\text{SFT*}}$, we leverage Bespoke-Stratos-17k\footnote{\url{https://huggingface.co/datasets/bespokelabs/Bespoke-Stratos-17k}}, which contains 17K tuples of questions, reasoning processes, and answers directly distilled from DeepSeek-R1~\cite{deepseekr1}. We also utilize two released CoT datasets to conduct supplementary experiments. The first is Sky-T1-data-17k\footnote{\url{https://github.com/NovaSky-AI/SkyThought}}, which is distilled from QwQ-32B-Preview—its reasoning abilities are reported to be weaker than those of DeepSeek-R1. The second is OpenThoughts-114k\footnote{\url{https://huggingface.co/datasets/open-thoughts/OpenThoughts-114k}}, which is distilled from DeepSeek-R1 and verified using a data curation recipe. We have chosen not to use some previously released CoT datasets (e.g., OpenLongCoT-SFT\footnote{\url{https://huggingface.co/datasets/SimpleBerry/OpenLongCoT-SFT}}) due to their significantly weaker reasoning abilities, while some benchmarks (e.g., AIME2024, OlympiadBench) are extremely challenging.

\begin{table}
\centering
\begin{tabular}{ll}
\toprule
\bf Dataset & \bf Size\\ 
\midrule
AIME2024 & 30\\
MATH-500 & 500\\
GSM8K & 1319\\
GPQA Diamond & 198\\
LiveCodeBench V2 & 511\\
MMLU & 14042\\
OlympiadBench (math-en) & 674\\
\bottomrule
\end{tabular}
\caption{Testing set statistics.}
\label{tab:stat}
\end{table}

\subsection{Experimental Details}
\label{appendix: Experimental Details}

In our work, we utilize Qwen2.5-7B-Instruct as the default model backbone and extend our evaluation to Llama3.1-8B-Instruct~\cite{DBLP:journals/corr/abs-2407-21783} and Mistral-7B-Instruct-v0.3~\cite{DBLP:journals/corr/abs-2310-06825}, along with other sizes of Qwen2.5 models, to validate the generalizability of our algorithm across diverse model architectures and sizes. We first establish a baseline by assessing the model’s zero-shot capabilities. Subsequent experiments leverage this result to quantify the performance improvements attributable to CRV and CogPO. During the CRV phase, the same generation hyperparameters are applied to the Critic, Rethinker, and Verifier for inference: temperature $T=0.7$, $\text{top\_p}=0.9$, and $\text{top\_k}=50$. The default backbone is DeepSeek-R1-Distill-Qwen-32B, while we test other backbone choices in the experiments. For CogPO training, the default $\beta$ values are: $\beta_\text{S}=0.1$, $\beta_\text{M}=0.2$, and $\beta_\text{L}=0.5$. Training details for all models and baselines are shown in Table~\ref{tab:train}.

On the Bespoke-Stratos-17k dataset: for the 3B model we use a single node with 8 A800 GPUs (80GB), with a training time of approximately 4 hours. For the 7B model, we use a single node with 8 A800 GPUs (80GB), with a training time of about 5 hours. For the 14B model, we use 4 nodes (each with 8 A800 GPUs), resulting in a training time of approximately 14 hours.

\begin{table}
\centering
\begin{small} 
\begin{tabular}{ll}
\toprule
\bf Hyperparameter & \bf Value\\ 
\midrule
\emph{CRV Stage} \\
\midrule
Batch size & 96\\
Learning rate & 1e-5\\
Learning epochs & 3.0\\
\midrule
\emph{CogPO Stage} \\
\midrule
Batch size & 96\\
Learning rate & 5e-7\\
Learning epochs & 1.0\\
\midrule
\emph{SFT (Baseline)} \\
\midrule
Batch size & 96\\
Learning rate & 1e-5\\
Learning epochs & 3.0\\
\midrule
\emph{DPO (Baseline)} \\
\midrule
Batch size & 96\\
Learning rate & 5e-7\\
Learning epochs & 1.0\\
$\beta$ & 0.1\\
\midrule
\emph{SimPO (Baseline)} \\
\midrule
Batch size & 96\\
Learning rate & 5e-7\\
Learning epochs & 1.0\\
$\beta$ & 2.0\\
$\gamma$ & 0.3\\
\bottomrule
\end{tabular}
\end{small}
\caption{Training hyperparameters.}
\label{tab:train}
\end{table}

\subsection{Results on Weaker CoT Dataset}

To demonstrate that our approach is truly superior to vanilla SFT over CoT datasets, we conduct an experiment on the Sky-T1 dataset, which is relatively weaker than Bespoke-Stratos-17k due to the choice of teacher model (i.e., QwQ-32B-Preview) and the data curation pipeline. The results are presented in Table~\ref{tab:sky}. As shown, in some cases the SFT baseline cannot even beat the zero-shot performance. This observation is consistent with their blog regarding model size and data quality\footnote{\url{https://novasky-ai.github.io/posts/sky-t1/}}. Nonetheless, by comparing our method with the SFT baseline, we observe clear improvements, which demonstrate the efficacy of our approach in enhancing the reasoning abilities of small models in various scenarios.

\begin{table}[t]
\centering
\begin{tabular}{l | lll}
\toprule
\bf Dataset/Model & \bf Zero-shot & \bf SFT & \bf Ours\\ 
\midrule
AIME2024 & 10.0 & \underline{16.7} & \bf 20.0\\
MATH-500 & \underline{73.6} & 73.2 & \bf 77.0\\
GPQA Diamond & \underline{33.3} & 28.8 & \bf 36.9\\
LiveCodeBench V2 & \underline{30.7} & 20.9 & \bf 33.3\\
\bottomrule
\end{tabular}
\caption{Performance comparison using Sky-T1-data-17k as the training set.}
\label{tab:sky}
\end{table}

\subsection{Results on Larger CoT Dataset}

We further evaluate the performance of our method using OpenThoughts-114k as the training set, which is much larger than other training sets. This dataset is distilled from DeepSeek-R1 and goes through several quality verification steps. The results are presented in Table~\ref{tab:openthoughts}. It can be seen that our method ultimately exhibits exceptionally strong reasoning performance, significantly surpassing SFT on all benchmarks. This underscores the scalability and generalizability of our CRV+CogPO framework to larger datasets.

\begin{table}[t]
\centering
\begin{tabular}{l | llll}
\toprule
\bf Dataset/Model & \bf Zero-shot & \bf SFT & \bf Ours\\ 
\midrule
AIME2024 & 10.0 & \underline{31.3} & \bf 43.3\\
MATH-500 & 73.6 & \underline{83.0} & \bf 88.4\\
GPQA Diamond & 33.3 & \underline{42.4} & \bf 42.9\\
LiveCodeBench V2 & 30.7 & \underline{39.9} & \bf 46.4\\
\bottomrule
\end{tabular}
\caption{Performance comparison using OpenThoughts-114k as the training set.}
\label{tab:openthoughts}
\end{table}

\subsection{Design Choice of the Critic}
\label{sec:criticAnalysis}

An initial, straightforward approach is to employ $\pi_{\text{base}}$ as the Critic. However, owing to the small model’s limited reasoning capability, it consistently faces difficulties in distinguishing the difficulty levels of CoTs effectively within our datasets. Note that for CoTs rated as ``easy'' or ``hard'', the CoT is either overly concise (omitting necessary steps) or excessively complex, rendering it unintelligible to the small model and preventing it from following the chain to arrive at the correct answer. Under these circumstances, it is clearly unreasonable to require the small model to classify the CoT difficulty that it cannot comprehend effectively.

Another intuitive CoT evaluation approach is to input the problem and its corresponding CoT into the small model and verify whether the model can arrive at the correct answer. However, applying this method directly would only partition CoT processes into ``correct'' or ``incorrect'' categories. For incorrect CoTs, this binary classification fails to distinguish the root cause of errors (i.e., whether the CoT is overly simplified or overly complex), which is critical for determining appropriate refinement strategies (e.g., expansion for overly simplified processes vs. simplification for overly complex ones).

Consequently, we utilize the larger and stronger LLM used in both the Rethinker and the Verifier (referred to as $\pi_{\text{large}}$) to act as the Critic. This involves guiding the large model to simulate the cognitive approach of the smaller model, $\pi_{\text{base}}$. The prompt template of the Critic is shown in Table~\ref{tab:prompt}. This setting is akin to educational practices, where professors, instead of students, customarily curate academic content across a spectrum of difficulty levels due to their broader knowledge base. As shown in Table~\ref{tab:critic}, the experiments clearly demonstrate the superior evaluative proficiency of the large model, confirming its advantage in categorizing CoT complexity from the perspective of the smaller model efficiently.

\section{Case Studies}

Case studies are presented in Tables~\ref{tab:case1} and~\ref{tab:case2}.

\section{Prompt Templates}
\label{sec:prompts}

Prompt templates for the Critic, Rethinker, and Verifier in our CRV system are shown in Table~\ref{tab:prompt}.

\begin{table*}
\centering
\begin{small}
\begin{tabular}{l|l}
\toprule
\bf Problem & Find the inverse of matrix $A = \begin{bmatrix}2 & 1 \\ 1 & 2\end{bmatrix}$\\ 
\midrule
\bf Answer & $A^{-1} = \frac{1}{3}\begin{bmatrix}2 & -1 \\ -1 & 2\end{bmatrix}$ \\
\midrule
\bf Original reasoning process & Calculate determinant $\det(A)=3$, thus $A^{-1}=\frac{1}{3}\begin{bmatrix}2 & -1 \\ -1 & 2\end{bmatrix}$ \\
(correct but unsuitable) &  \\
\midrule
\bf Extended reasoning process & Compute determinant: $2 \times 2 - 1 \times 1 = 3$ \\
(correct and suitable)& Construct adjugate: $\begin{bmatrix}2 & -1 \\ -1 & 2\end{bmatrix}$ \\
& Normalize: $ A^{-1} =\frac{1}{3}\begin{bmatrix}2 & -1 \\ -1 & 2\end{bmatrix}$ \\
\midrule
\bf Incorrect reasoning process & Swap diagonal elements: $\begin{bmatrix}2 & 1 \\ 1 & 2\end{bmatrix} \rightarrow \begin{bmatrix}2 & -1 \\ -1 & 2\end{bmatrix}$ \\
\bottomrule
\end{tabular}
\end{small}
\caption{Case study of how the reasoning process is extended.}
\label{tab:case1}
\end{table*}

\begin{table*}
\centering
\begin{small}
\begin{tabular}{l|l}
\toprule
\bf Problem & Find the area of a triangle with vertices at $(0,0)$, $(3,0)$, and $(0,4)$\\ 
\midrule
\bf Answer & $6$ \\
\midrule
\bf Original reasoning process & Vector Representation: $\vec{AB} = \begin{pmatrix}3\\0\end{pmatrix}, \vec{AC} = \begin{pmatrix}0\\4\end{pmatrix}$ \\
(correct but unsuitable) & Determinant Method: $\text{Area} = \frac{1}{2}\left|
        \begin{vmatrix}
        3 & 0 \\
        0 & 4
        \end{vmatrix}
        \right| = \frac{1}{2}(12) = 6$ \\
\midrule
\bf Simplified reasoning process & Recognize right-angled triangle: $\text{Base} = 3,\ \text{Height} = 4$ \\
(correct and suitable)&Apply elementary formula: $\text{Area} = \frac{1}{2} \times \text{Base} \times \text{Height} = \frac{1}{2} \times 3 \times 4 = 6$ \\
\midrule
\bf Incorrect reasoning process & $\text{Area} = \frac{1}{2}(\text{Sum of sides}) = \frac{1}{2}(3+4+5) = 6$ \\
\bottomrule
\end{tabular}
\end{small}
\caption{Case study of how the reasoning process is simplified.}
\label{tab:case2}
\end{table*}

\begin{table*}
\centering
\begin{small}
\begin{tabular}{l|l}
\toprule
\bf Role & \bf Prompt Template\\ 
\midrule
\bf Critic & You are a highly capable evaluator. \\
& Your task is to assess the given reasoning process from the perspective of a small language model (e.g., \\
& 7B). \\
& Specifically, determine whether the reasoning process provides sufficient detail for a small model to  \\
& solve the problem, or whether it is too terse (i.e., lacking critical details) or too complex (i.e., containing \\
& unnecessary or confusing steps). \\
& Complexity Definitions (from the perspective of a small model): \\
& - Easy: The reasoning process is overly terse; it omits essential details that a small model needs to solve\\
&  the problem. \\
& - Medium: The reasoning process is appropriately balanced, offering enough detailed guidance. \\
& - Hard: The reasoning process is overly complex, with extraneous or convoluted steps that could hinder \\
&  a small model to follow it. \\
& Output Format: \\
& You must output exactly one word: easy, medium, or hard.  \\
\midrule
\bf Rethinker & You are a helpful assistant who is highly skilled at extending reasoning processes. \\
(easy)& Given a problem ,its correct answer and its terse reasoning process, your task is to extend the reasoning \\
& process by adding necessary details and intermediate steps so that a small language model (e.g., a 7B \\
& model) can follow the extended reasoning process to solve the problem. \\
& You should add necessary steps and details based on the correct answer.\\
& You must output ONLY the extended reasoning process with no additional explanation or commentary. \\
\midrule
\bf Rethinker & You are a helpful assistant who is highly skilled at simplifying reasoning processes. \\
(hard)& Given a problem, its correct answer and its overly complex reasoning process, your task is to simplify the \\
& reasoning process so that a small language model (e.g., a 7B model) can reliably follow the steps to solve \\
& the problem. \\
& You should remove redundancies or use simpler method on the basis of correct \\
& answer.\\
& You must output ONLY the simplified reasoning process with no additional explanation or commentary. \\
\midrule
\bf Verifier & You are a highly capable Verifier. \\
& Your task is to assess a given reasoning process based on a problem and its correct answer. \\
& Specifically, determine whether the reasoning process is sufficient and accurate for you to reach the correct \\
& answer. \\
& If the reasoning process correctly guides you to derive the the correct answer, output YES. \\
& If the reasoning process fails to guide you to the correct answer, output NO. \\
& You must output exactly one word: YES or NO.  \\
\midrule
\bf Rethinker & You are an assistant who is skilled at converting correct reasoning processes to incorrect reasoning \\
(incorrect)&  processes. Given a problem, its answer and its correct reasoning process, your task is to corrupt the correct \\
 & reasoning process by introducing logical fallacies and misleading steps, so that a small language model (e.g., \\
& a 7B model) cannot follow the incorrect reasoning process to solve the problem. \\
& You must output ONLY the incorrect reasoning process with no additional explanation or commentary. \\
\bottomrule
\end{tabular}
\end{small}
\caption{Prompt templates for the CRV+CogPO framework.}
\label{tab:prompt}
\end{table*}

\end{document}